\documentclass[letterpaper, 10 pt, conference]{ieeeconf}  
\usepackage{diagbox}

\usepackage{epsfig} % for postscript graphics files
\usepackage{amsmath} % assumes amsmath package installed
\usepackage{amssymb}  % assumes amsmath package installed
\usepackage{color}
\usepackage[linesnumbered,ruled]{algorithm2e}
\usepackage[normalem]{ulem} %to strike the words
\makeatletter
\let\NAT@parse\undefined
\makeatother
\usepackage{hyperref}
\hypersetup{pdfstartview=FitH,
            colorlinks=true,
            linkcolor=red,
            anchorcolor=blue,
            citecolor=green
            }
\usepackage[table]{xcolor}
\usepackage{cite}
\usepackage{booktabs}
\usepackage{multirow}
\usepackage[caption=false,font=footnotesize]{subfig}

\newcounter{RNum}
\renewcommand{\theRNum}{\arabic{RNum}}
\usepackage{pifont}
\usepackage{booktabs}
\newcommand{\cmark}{\ding{51}}
\newcommand{\Remark}{\noindent\textit{\textbf{Remark}~\refstepcounter{RNum}\textbf{\theRNum}: }}
\usepackage{xcolor}
\definecolor{table_c}{RGB}{240,240,240}
\definecolor{figure_c}{RGB}{241,241,241}
\IEEEoverridecommandlockouts                              
\title{\LARGE \bf
Prompt-Driven Temporal Domain Adaptation \\ for Nighttime UAV Tracking
}

\author{Changhong Fu$^{1*}$, Yiheng Wang$^{1}$, Liangliang Yao$^{1}$, Guangze Zheng$^{2}$, Haobo Zuo$^{2}$, and Jia Pan$^{2}$ % <-this % stops a space
\thanks{*Corresponding author}% <-this % stops a space
\thanks{$^{1}$C. Fu, Y. Wang, and L. Yao are with the School of Mechanical Engineering, Tongji University, Shanghai 201804, China. 
\itshape{Email: changhongfu@tongji.edu.cn}}%
\thanks{$^{2}$G. Zheng, H. Zuo, and J. Pan are with the Department of Computer Science, the University of Hong Kong, Hong Kong, China.}%
}

\begin{document}
\maketitle
\thispagestyle{empty}
\pagestyle{empty}

%%%%%%%%%%%%%%%%%%%%%%%%%%%%%%%%%%%%%%%%%%%%%%%%%%%%%%%%%%%
%%%%%%%%%%%%%%%%%%%%%%%%%%%%%%%%%%%%%%%%%%%%%%%%%%%%%%%%%%%
\begin{abstract}
Nighttime UAV tracking under low-illuminated scenarios has achieved great progress by domain adaptation (DA). However, previous DA training-based works are deficient in narrowing the discrepancy of temporal contexts for UAV trackers. To address the issue, this work proposes a prompt-driven temporal domain adaptation training framework to fully utilize temporal contexts for challenging nighttime UAV tracking, \textit{i.e.}, TDA. Specifically, the proposed framework aligns the distribution of temporal contexts from daytime and nighttime domains by training the temporal feature generator against the discriminator. The temporal-consistent discriminator progressively extracts shared domain-specific features to generate coherent domain discrimination results in the time series. Additionally, to obtain high-quality training samples, a prompt-driven object miner is employed to precisely locate objects in unannotated nighttime videos. Moreover, a new benchmark for long-term nighttime UAV tracking is constructed. Exhaustive evaluations on both public and self-constructed nighttime benchmarks demonstrate the remarkable performance of the tracker trained in TDA framework, \textit{i.e.}, TDA-Track. Real-world tests at nighttime also show its practicality. The code and demo videos are available at \url{https: //github.com/vision4robotics/TDA-Track}.
\end{abstract}

%%%%%%%%%%%%%%%%%%%%%%%%%%%%%%%%%%%%%%%%%%%%%%%%%%%%%%%%%%%
%%%%%%%%%%%%%%%%%%%%%%%%%%%%%%%%%%%%%%%%%%%%%%%%%%%%%%%%%%%
\section{Introduction}

\indent Visual object tracking on intelligent unmanned aerial vehicles (UAVs) has caught widespread interest for its versatility in various real-world applications, \textit{e.g.}, navigation~\cite{xiao2017uav}, search and rescue mission~\cite{mittal2019vision}, as well as
security and surveillance~\cite{bozcan2020uav}.
%sandino2022reducing,
%Given the object in the initial frame, trackers determine its position in the upcoming frames. 
Significant advancements in UAV tracking performance have been made in favorably-illuminated scenarios~\cite{bertinetto2016siamfc,li2018siamrpn,li2019siamrpn++,xu2020siamfc++}. However, the images captured by UAVs at night have much lower contrast, brightness, and signal-to-noise ratio~\cite{ye2022unsupervised} than ones captured in the daytime, resulting in a huge domain discrepancy between nighttime and daytime visual feature spaces. Hence, state-of-the-art (SOTA) trackers suffer from severe tracking capability degradation. Moreover, the temporal contexts, vital information contained in the consecutive frames, haven't been utilized in nighttime UAV tracking. Consequently, robust nighttime UAV tracking is far from being properly addressed.\\
\begin{figure}[t]
    \centering
    \includegraphics[width=1\linewidth]{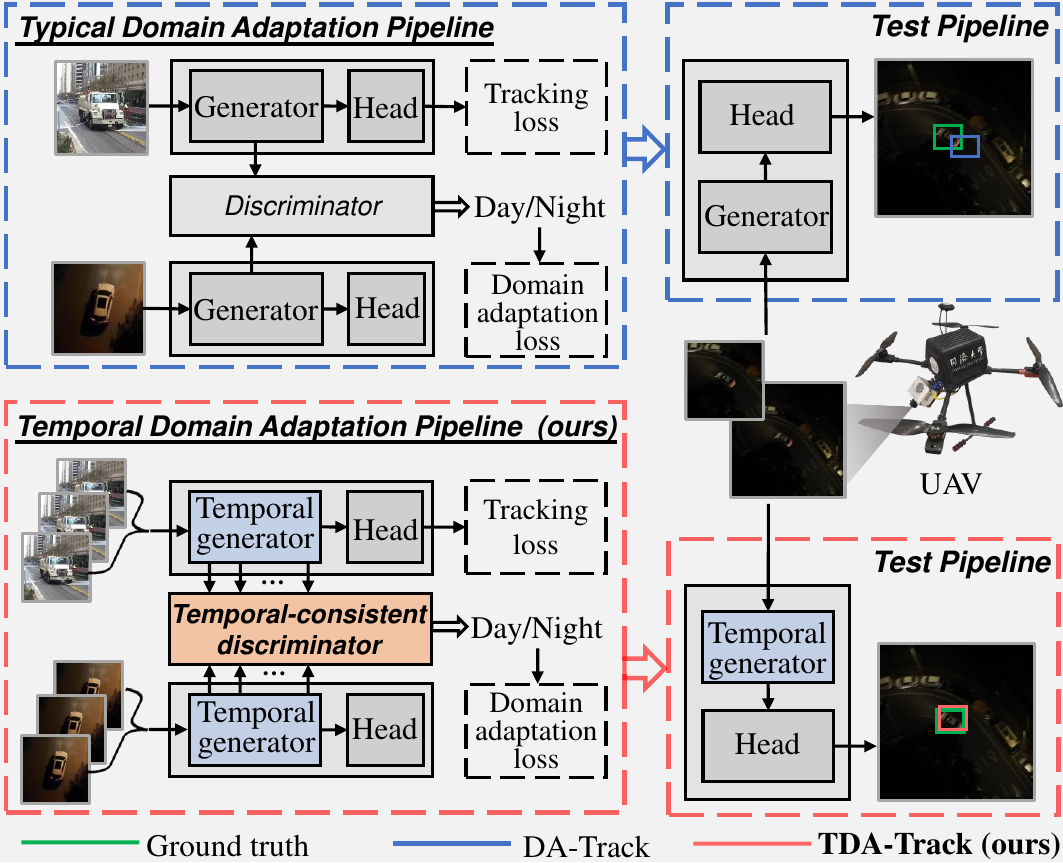}
    \caption{Comparison with previous domain adaptation (DA) training framework for nighttime UAV tracking. The proposed temporal domain adaptation (TDA) training framework generates the temporal contexts among daytime and nighttime images, and then narrows the feature discrepancy of temporal contexts from different domains with the temporal-consistent discriminator. (Image frames are from GOT-10k~\cite{huang2019got} and NAT2021-\textit{train}~\cite{ye2022unsupervised}.)}
    \vspace{-20pt}
    \label{fig:fig1 comparison}
\end{figure}
\indent One promising solution to the domain discrepancy challenge encountered by UAV tracking is domain adaptation (DA)~\cite{ye2022unsupervised, fu2024SAMDA}. Specifically, the DA training framework includes a feature generator and a discriminator. Trained to deceive the discriminator, the generator learns to extract domain-invariant features from unordered daytime and nighttime training samples. Meanwhile, the classification and localization capabilities are trained under the supervision of daytime annotations. However, the DA training on static image pairs neglects the strong temporal contexts that inherently exist among consecutive frames, which is unsatisfactory for UAV tracking which highlights the continuity. The feature distribution is not consistently aligned among consecutive frames, leading to the failure in long-term object tracking.  Moreover, the plain Transformer discriminators~\cite{ye2022unsupervised, fu2024SAMDA} fail to generate coherent discrimination results due to context changes among consecutive frames, which undermines the practicality of the DA framework for UAV tracking. \textbf{\textit{Therefore, how to consistently align temporal contexts and to obtain coherent discrimination is worth attention in DA training-based nighttime UAV tracking}}.\\
 \indent Previous DA training frameworks for nighttime UAV tracking~\cite{ye2022unsupervised, fu2024SAMDA} discover nighttime training samples with video saliency detection model~\cite{zhang2021dynamic} and segment anything model (SAM)~\cite{kirillov2023segment}. However, these object discovery approaches focus on the object position in the single frame and are unsatisfactory in building smooth trajectories of uncovered objects, which results in inconsistent training samples and suboptimal performances. \textbf{\textit{Hence, how to generate high-continuity training samples for nighttime UAV tracking remains unresolved.}}\\
\indent This work proposes a novel framework, \textit{i.e.}, TDA, which explores temporal contexts in DA training for nighttime UAV tracking and utilizes prompts to obtain high-quality training samples. 
The Baseline tracker~\cite{cao2022tctrack} trained in TDA framework is symbolized as TDA-Track. 
As illustrated in Fig.~\ref{fig:fig1 comparison}, to alleviate the inconsistency of temporal contexts, the proposed training framework aligns the distribution of temporal contexts and image features from daytime and nighttime domains. Specifically, given a sequence with \(t\) frames, the temporal generator maps frames into temporal contexts and image features, which are classified into different domains by discriminator. In the adversarial training manner, the temporal generator is trained to deceive the discriminator and consistently obtain domain-invariant representations across the time series. To improve domain adaptability, a novel temporal-consistent discriminator is designed to make more coherent discrimination results with common features, which are robust representations progressively extracted from temporal contexts. 
Moreover, benefited from progress in multi-modal learning~\cite{radford2021learning,li2022grounded}, a prompt-driven object mining approach is proposed to acquire high-quality training samples. Objects are mined by detection~\cite{liu2023grounding} with text prompt and associated into smooth trajectories in the time series. The main contributions of this work are summarized as follows:
\begin{itemize}
\item A novel temporal domain adaptation training framework is proposed for nighttime UAV tracking. To the best of our knowledge, our method is the first work to explore the power of temporal contexts in training nighttime UAV trackers.
\item An innovative temporal-consistent Transformer discriminator is designed to boost the coherency of discrimination results by progressively extracting common features from temporal contexts. 
\item A new prompt-driven object mining approach is put forward to mine highly-continuous training patches from nighttime videos. It surpasses previous pre-processing methods by excellent consistency and valuable object categories.
\item A novel long-term nighttime UAV tracking benchmark is constructed to enlarge the existing benchmarks for a fair comparison. Evaluations on both public and self-constructed benchmarks show the remarkable nighttime UAV tracking performance of TDA-Track. Real-world deployment on the UAV platform demonstrates its robustness and practicality.
\end{itemize}

%%%%%%%%%%%%%%%%%%%%%%%%%%%%%%%%%%%%%%%%%%%%%%%%%%%%%%%%%%%
%%%%%%%%%%%%%%%%%%%%%%%%%%%%%%%%%%%%%%%%%%%%%%%%%%%%%%%%%%%
\section{Related Works}
\subsection{Nighttime UAV Tracking}
Real-world UAV tracking applications are severely impeded by low illumination at nighttime. Recently, the approaches to boost nighttime UAV tracking performance are founded on either low-light enhancement or DA.
For enhancement-based nighttime UAV tracking~\cite{ye2021darklighter,ye2022tracker}, different enhancers are proposed to improve the image illumination ahead of trackers. However, due to the limited relationship between low-light image enhancement and UAV tracking, integrating enhancers and trackers in the plug-and-play manner leads to suboptimal performance.
For DA training-based nighttime UAV tracking~\cite{ye2022unsupervised, fu2024SAMDA}, trackers are trained on static image pairs to obtain domain-invariant features for predicting the object location in upcoming frames. However, existing DA training frameworks are insufficient in aligning the distribution of temporal contexts, restricting performance improvement of nighttime UAV tracking. 
% Moreover, previous frameworks sample many training patches that contain no target, which hinders the training efficiency.

\subsection{Temporal Contexts}
Temporal contexts have aroused much research interest for the effectiveness in object tracking. B. Yan~\textit{et al.}~\cite{yan2021learning} explicitly replace the template with high-confidence patches. Z. Fu~\textit{et al.}~\cite{fu2021stmtrack} and N. Wang~\textit{et al.}~\cite{wang2021transformer} design architectures to fuse previous template features into robust template representations. To exploit temporal contexts more comprehensively, Z. Cao~{\textit{et al.}}~\cite{cao2022tctrack, cao2023towards} introduce temporal knowledge into feature extraction and similarity map refinement. 
In challenging nighttime UAV tracking, temporal contexts among consecutive frames are valuable hints for the possible object location. However, existing DA methods for nighttime UAV tracking have not taken temporal contexts into consideration. Moreover, the lack of high-quality training samples that follow objects with smooth trajectories poses great challenges for trackers to learn temporal contexts.

%直接讲domain adaptation即可。在II. B中，首先指出DA目前主要是静态的，但在da-based tracking任务中重要的时序性没有获得足够重视，近来的[18-21]展现了时序对tracking的重要意义。（一笔带过即可）。时序性DA能够保证在时间维度上对图像视觉建模对目标域映射的连贯性，从而能够保证作为时序任务的da-based tracking的跟踪表现，因而成为一项亟待解决的issue
\subsection{Temporal Domain Adaptation}
To achieve favorable performance on the target domain with the power of temporal contexts, mappings from source domain representations to target domain temporal contexts should be coherent in the time series, which remains a problem in temporal domain adaptation.
To alleviate the problem, B. Pan~\textit{et al.}~\cite{pan2020adversarial} design a discriminator architecture for domain distribution matching and temporally aligned distribution matching. I. Shin~\textit{et al.}~\cite{shin2021unsupervised} adopt a sequence discriminator to take soft segmentation maps in a sequence for domain classification. X. Feng~\textit{et al.}~\cite{Feng_2021} and D. Guan~\textit{et al.}~\cite{guan2021domain} design two discriminators to classify spatial features from a single image and spatial-temporal features from consecutive frames.
However, these works adopt simple discriminator architectures for temporal domain adaptation. The incoherent discrimination results misguide the generator in adversarial training and  undermines the domain adaptability. Moreover, temporal domain adaptation for nighttime UAV tracking has not been investigated yet.

%%%%%%%%%%%%%%%%%%%%%%%%%%%%%%%%%%%%%%%%%%%%%%%%%%%%%%%%%%%
%%%%%%%%%%%%%%%%%%%%%%%%%%%%%%%%%%%%%%%%%%%%%%%%%%%%%%%%%%%
\section{Proposed Method}
\begin{figure*}[h]
    \centering
    \includegraphics[width=1\linewidth]{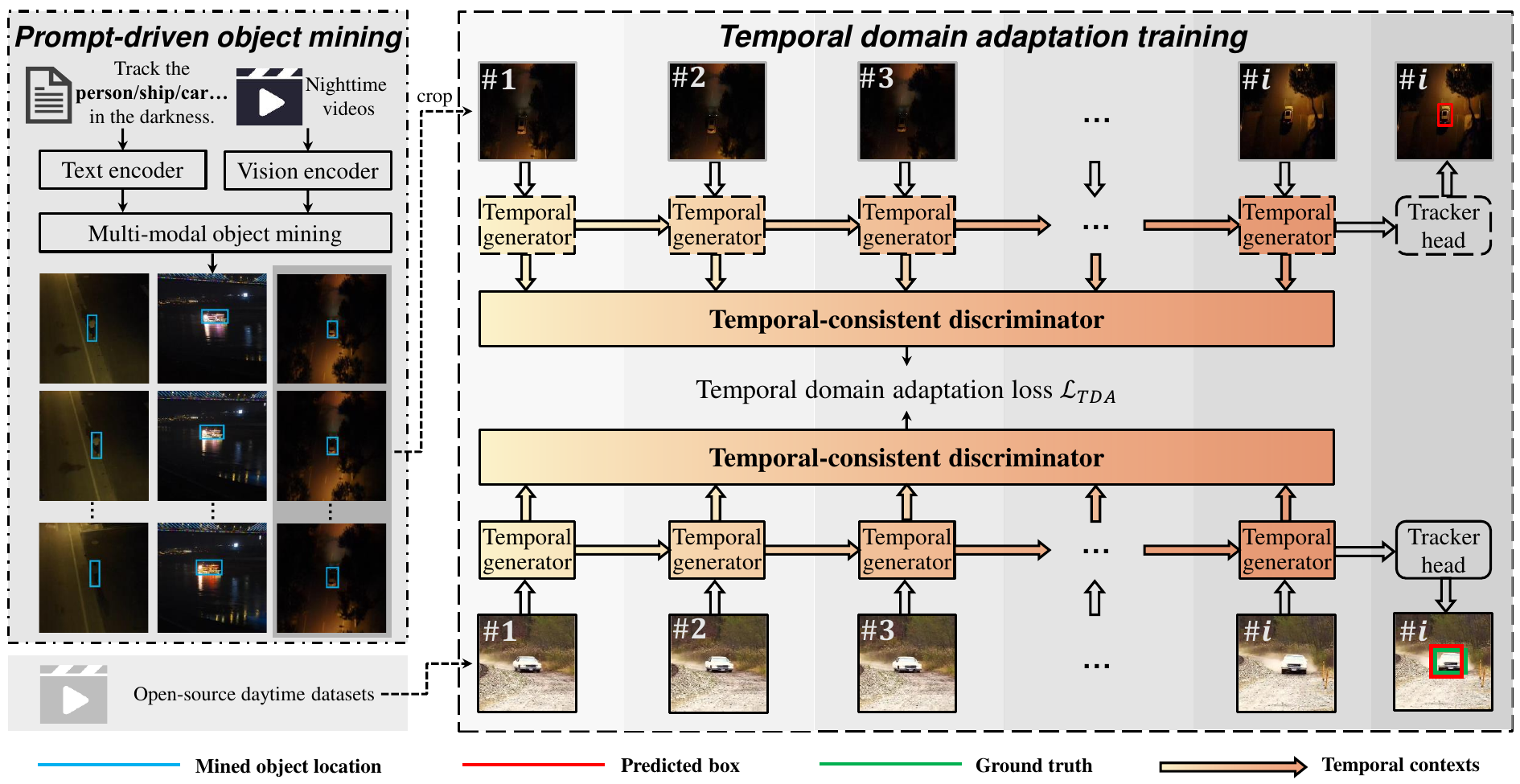}
    %\captionsetup{skip=-5pt}
    \caption{Overview of the temporal day-to-night domain adaptation framework for nighttime UAV tracking. The \textit{temporal generator} learns to generate temporal contexts that are more adaptive to the nighttime domain. The \textit{temporal-consistent discriminator} is trained to classify features and temporal contexts into different domains based on progressively extracted domain-specific representations. Prompt-driven object mining locates valuable objects with text prompts and builds their smooth trajectories in the time series. (Image frames are from GOT-10k~\cite{huang2019got} and NAT2021-\textit{train}~\cite{ye2022unsupervised}.)}
    \vspace{-5pt}
    \label{fig:fig2 overall architecture}
\end{figure*}

The overview of the TDA training pipeline is shown in Fig.~\ref{fig:fig2 overall architecture}. The proposed framework innovatively introduces temporal contexts into domain adaptation from two perspectives: 1) To boost nighttime UAV tracking with robust temporal representations, the temporal contexts of consecutive frames from different domains are aligned jointly. 2) The discriminator progressively extracts common features from temporal contexts, which benefits domain classification by highly representative features for nighttime attributes. Moreover, to obtain high-quality training samples, a prompt-driven object mining approach is provided to locate objects from unlabelled videos and build smooth trajectories.

\subsection{Temporal Domain Adaptation Training}
\begin{figure*}[h]
    \centering
    \vspace{5pt}
    \includegraphics[width=1\linewidth]{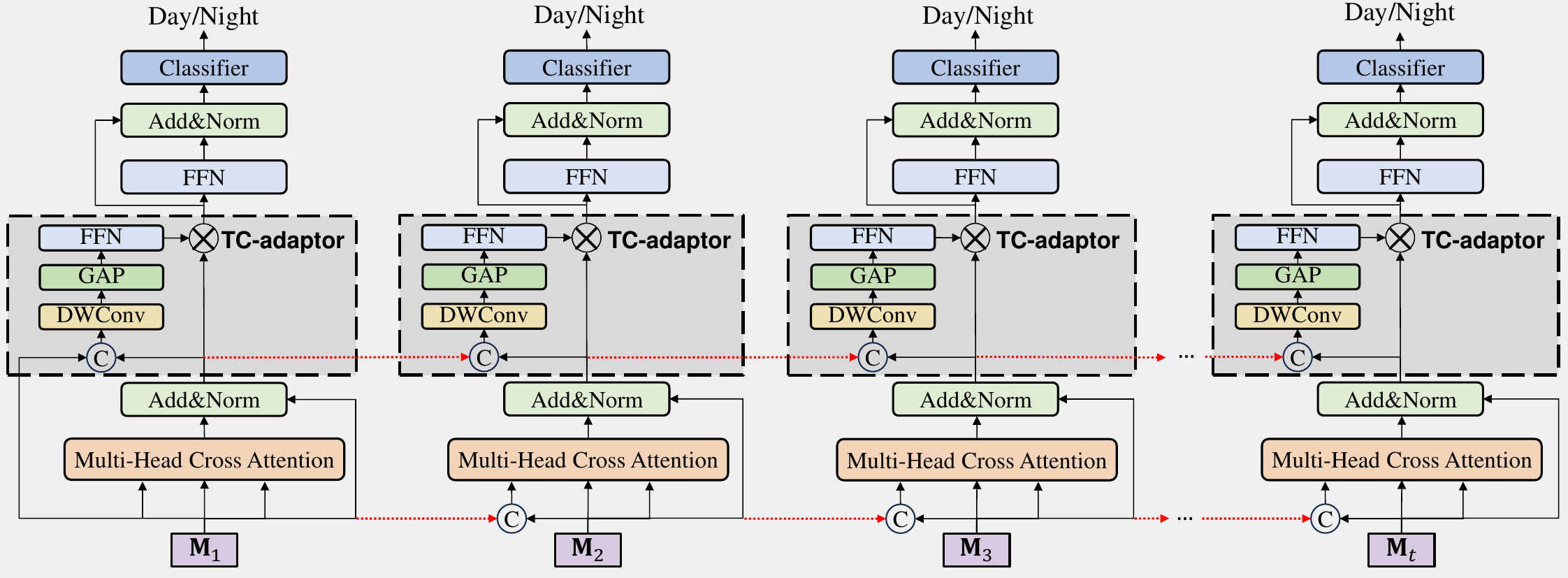}
    \caption{The structure of the temporal-consistent discriminator. \(\mathrm{\mathbf{M_{\textit{i}}}}\) denotes the temporal contexts encoded from the first \(i\) frames. The utilization of temporal contexts is marked with \textcolor{red}{red} dotted lines. Better representations oriented for daytime or nighttime attributes are progressively extracted, which enables more robust discrimination.}
    \label{fig:fig3 discriminator}
    \vspace{-15pt}
\end{figure*}
The temporal domain adaptation training framework (TDA) consists of a temporal tracker and a domain discriminator. Generally, a temporal tracker includes a temporal generator and tracker head. The temporal feature generation can be described as:
\begin{equation}
\mathrm{\mathbf{F}}_t= \varphi(I_t,\mathrm{\mathbf{M}})~,
\end{equation}
where \(\mathrm{\mathbf{F}}_t\) denotes the temporal features generated from current frame \(I_t\) and temporal contexts \(\mathrm{\mathbf{M}}\in \mathbb{R}^{ (t-1) \times C}\) by feature extraction and temporal modelling function \(\varphi\).\\  
\indent For the huge discrepancy between daytime and nighttime domains, the temporal generator trained on well-illuminated datasets can hardly extract satisfactory temporal representations from nighttime images captured by UAVs. Previous DA methods~\cite{ye2022unsupervised,fu2024SAMDA} narrow feature distribution discrepancy of \(\mathrm{\mathbf{F}}_t\). However, inappropriate temporal contexts are likely to misguide temporal generators into generating unsatisfactory representations of upcoming frames, resulting in tracking failure. Consequently, the proposed framework aims to take full advantage of robust temporal contexts by aligning their distribution. The loss functions of the discriminator and tracker, denoted as \(E_{D}\) and \(E_{G}\), are formulated as follows:
\begin{equation}
\begin{aligned}
E_{D}&=\mathcal{L}_{D}(\theta_1(\mathrm{\mathbf{F}}_t),l_{t})+\sum_{i=1}^{t-1} \mathcal{L}_{D}(\theta_2(\mathrm{\mathbf{M}}_i),l_{t})~,\\
E_{G}&=\mathcal{L}_{gt}+\mathcal{L}_{G}(\theta_1(\mathrm{\mathbf{F}}_t),l_{f})+\sum_{i=1}^{t-1} \mathcal{L}_{G}(\theta_2(\mathrm{\mathbf{M}}_i),l_{f})~,
\end{aligned}
\end{equation}
where \(\theta_{1}\), \(\theta_{2}\) denote the discriminators for temporal features and temporal contexts, and \(l_{t}\), \(l_{f}\) represent the true or false bool label for whether the domain classification is correct. \(\mathcal{L}_{D}\), \(\mathcal{L}_{G}\), and \(\mathcal{L}_{gt}\) respectively denote the loss for domain discrimination, feature alignment, as well as classification and regression.\\
\indent The adversarial training process is formulated as:
\begin{equation}
\begin{aligned}
    (\hat{\theta}_{1}, \hat{\theta}_{2}) &= \underset{\theta_{1}, \theta_{2}}{min} \, E_{D}(\hat{\varphi}, \theta_{1}, \theta_{2})~,\\
    (\hat{\varphi}, \hat{\psi}) &= \underset{\varphi, \psi}{min} \, E_{G}(\varphi, \psi, \hat{\theta}_{1}, \hat{\theta}_{2})~,
\end{aligned}
\end{equation}
where \(\psi\) denotes the tracker head. \(\hat{\varphi}\), \(\hat{\psi}\) represent the temporal generator and tracker head with learned parameters, while \(\hat{\theta}_{1}\), \(\hat{\theta}_{2}\) denote the learned discriminators. In cases where \(\mathrm{\mathbf{F}}_t\) and \(\mathrm{\mathbf{M}}\) are mapped into shared space by the same temporal generator, only one discriminator is necessary.\\
\indent With contradictory training objectives, the temporal tracker and the discriminator gradually reach convergence. The reduction of discrepancy in both temporal contexts and image features guarantees robust representations vital for long-term nighttime UAV tracking. \\
\Remark While previous DA frameworks overlook the discrepancy of temporal contexts, the proposed novel framework narrows the distribution gap of temporal contexts from different domains. Aligned representations for both image features and temporal contexts enable TDA-Track to perform robustly in poorly illuminated nighttime scenarios. 

\subsection{Temporal-Consistent Discriminator}
\indent The precision and robustness of the discriminator substantially contribute to the effectiveness of adversarial training. However, discrimination is interfered by inaccurate features extracted from a single frame with noises, resulting in incoherent discrimination results in the time series. To alleviate the problem, shared features among consecutive frames are more reliable grounds for domain discrimination, since they tend to be more robust and representative for domain-specific distributions. Hence, a temporal-consistent discriminator is designed for the proposed TDA framework, as presented in Fig.~\ref{fig:fig3 discriminator}. It progressively extracts and refines common features from temporal contexts and image features with the cross attention mechanism and a novel adaptor, to boost the accuracy and coherence of domain discrimination.\\
\indent Specifically, multi-head cross attention first extracts the common features from consecutive temporal contexts. Given temporal contexts \(\mathrm{\mathbf{M}}_{i-1}\) and \(\mathrm{\mathbf{M}}_{i}\), they are concatenated as input for query projection. The cross attention mechanism can be formulated as:
\begin{equation}
\mathrm{\mathbf{M}}_{i}^{'} = \mathrm{Norm}(\mathrm{\mathbf{M}}_{i} + \mathrm{Attn}((\mathrm{\mathbf{M}}_{i-1},\mathrm{\mathbf{M}}_i), \mathrm{\mathbf{M}}_{i}, \mathrm{\mathbf{M}}_{i}))~,
\end{equation}
where \(\mathrm{Norm}\) represents the layer normalization, while \(\mathrm{Attn}\) denotes the multi-head cross attention. \(\mathrm{\mathbf{M}}_{i}^{'}\) denotes the extracted common features.\\
\indent It's noted that when encountered with severe nighttime UAV tracking challenges, the features change greatly across the time series, resulting in the inaccuracy of the common features extracted. Hence, we propose a temporal-consistent adaptor (TC-adaptor) to refine the common features by concentrating on channels insensitive to context changes. \\
\indent This work concatenates (Concat) the features of two consecutive image frames, and utilizes a depthwise separable convolution layer (DWConv) to discover the differences. The extent of feature difference is obtained by global average pooling (GAP) to compress the size of features, \textit{i.e.}, \(\mathrm{\mathbf{D}}_{i}=\mathrm{GAP}(\mathrm{DWConv}(\mathrm{Concat}(\mathrm{\mathbf{M}}_{i-1}^{'}, \mathrm{\mathbf{M}}_{i}^{'})))\). The feed-forward network (\(\mathrm{FFN}\)) and multiplication follow to enhance the feature robustness:
\begin{equation}
\mathrm{\mathbf{M}}_{i}^{''} = \mathrm{\mathbf{M}}_{i}^{'}*\mathrm{FFN}(\mathrm{\mathbf{D}}_{i})~,
\end{equation}
where \(\mathrm{\mathbf{M_{\textit{i}}^{''}}}\) is the refined common representations. \\
\indent Then, a linear classifier tells which domain the extracted common features belong to. \\
\Remark The temporal-consistent discriminator progressively encodes domain-specific representation and denoises irrelevant information, which helps narrow the domain discrepancy of temporal contexts and image features.

\begin{figure*}[!t]
    \centering
    \vspace{5pt}
    \colorbox{figure_c}{
        \includegraphics[width=0.95\linewidth]{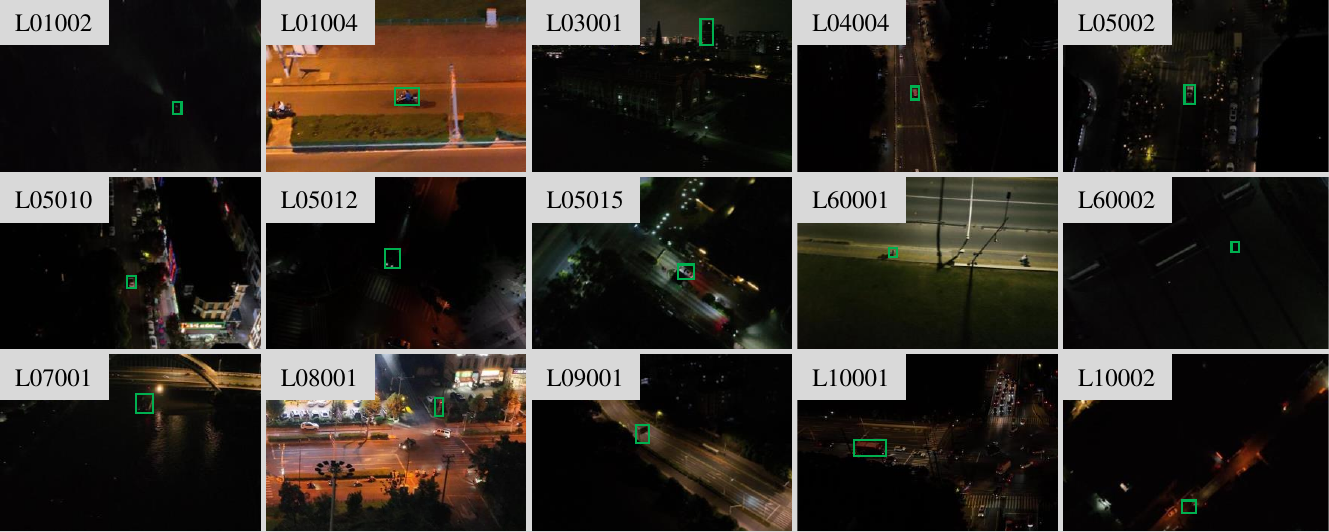}
        }
     \caption{The first frames of typical scenes in NAT2024-1. The tracking objects are marked with green boxes. The dark environments pose a great challenge to nighttime UAV tracking.}
    \label{fig:dataset}
    \vspace{-15pt}
\end{figure*}

\subsection{Prompt-Driven Object Mining}
To improve the coherence of cropped training patches for better temporal tracking performance, this work develops a two-stage strategy to alleviate the problem: 1) utilize multi-modal object detection to find object candidates specified by text prompt in each frame. 2) build smooth trajectories with tracking by detection method.\\
\indent Given a (\textit{Video, Text}) pair, text features and visual features are extracted by a text encoder and an image encoder, respectively:
\begin{equation}
(\mathrm{\mathbf{E}}^{v}_{i},~\mathrm{\mathbf{E}}^{t}) =( \chi(Image_{i}),~\phi(Text))~,
\end{equation}
where \(\mathrm{\mathbf{E}}_{i}^{v}\) denotes the visual features of the \(i_{th}\) frame \(Image_{i}\) extracted by the image  encoder \(\chi\), while \(\mathrm{\mathbf{E}}^{t}\) denotes the text features of the given \(Text\) extracted by the text encoder \(\phi\).\\
\indent Then text features and visual features are fused within a deep network to strengthen their correlation in vision-text space. A predictor follows to output pairs of object boxes and category names. The object boxes in the time series are associated into smooth trajectories~\(Trajs\). The process can be formulated as:
\begin{equation}
Trajs=\mathrm{Associate}(\vartheta(\mathrm{\mathbf{E}}^{v},\mathrm{\mathbf{E}}^{t}))~, 
\end{equation}
where \(\vartheta\) represents the deep vision-text fusion and prediction~\cite{liu2023grounding}, while \(\mathrm{Associate}(\cdot)\) denotes the association function~\cite{nettrack}.\\
\indent High-quality training patches are cropped based on \(Trajs\):
\begin{equation}
\begin{aligned}
[\mathrm{Z}^t_{1},\dots,\mathrm{Z}^t_{n}] &= \mathrm{ZCrop}(I_t; [\mathrm{B}^t_1,\dots, \mathrm{B}^t_n])~,\\
[\mathrm{X}^t_{1},\dots,\mathrm{X}^t_{n}] &= \mathrm{XCrop}(I_t; [\mathrm{B}^t_1,\dots, \mathrm{B}^t_n])~,
\end{aligned}
\end{equation}
where \(\mathrm{B}^t_{i}\) denotes the predicted bounding box of \(i_{th}\) trajectory in \(t_{th}\) frame \(I_{t}\). \(\mathrm{ZCrop}\), \(\mathrm{XCrop}\) denote cropping images into different sizes with object bounding boxes in the patch center. \(\mathrm{Z}^t_{i}\) and \(\mathrm{X}^t_{i}\) are respectively the cropped search patch and template patch corresponding to \(\mathrm{B}^t_{i}\).\\
\Remark The prompt-driven object mining approach generalizes well in the nighttime scenarios and builds smooth trajectories from temporal information, which generates high-continuity samples for the temporal domain adaptive training of nighttime UAV trackers.

%%%%%%%%%%%%%%%%%%%%%%%%%%%%%%%%%%%%%%%%%%%%%%%%%%%%%%%%%%%
%%%%%%%%%%%%%%%%%%%%%%%%%%%%%%%%%%%%%%%%%%%%%%%%%%%%%%%%%%%
\section{Novel Benchmark: NAT2024-1}
A large-scale long-term nighttime UAV tracking benchmark, \textit{i.e.}, NAT2024-1, is constructed to realistically evaluate the tracking performance and application value of tracking models. As illustrated in TABLE~\ref{tab:dataset}, the benchmark consists of 40 long-term image sequences with over 70K frames in total. The sequences are newly captured in diverse scenes by a DJI Mavic 3 Classic UAV. The benchmark includes various typical UAV tracking targets, \textit{e.g.}, electric bikes, pedestrians, and vehicles. 
Several typical nighttime UAV tracking cases are shown in Fig.~\ref{fig:dataset}. \\
\indent Five attributes are annotated, \textit{i.e.}, aspect ratio change (ARC), fast motion (FM), illumination variation (IV), low ambient illumination (LAI), and scale variation (SV). It's worth noting that 35 sequences in NAT2024-1 feature as low ambient intensity, as shown in TABLE~\ref{tab:dataset}, which are representative scenes in nighttime UAV tracking.

\begin{table}[b]
  \centering
  \caption{Comparison of NAT2024-1 with long-term subsets of existing nighttime UAV tracking benchmarks.}
  \vspace{-5pt}
  \colorbox{table_c}{
  %\resizebox{0.95\linewidth}{!}{
    %\setlength{\heavyrulewidth}{1.2pt}
    \begin{tabular}{lcccc}
    \toprule
    Benchmarks & Long Sequences & Frames & LAI attribute \\
    \midrule
    NAT2021-\textit{L-test}~\cite{ye2022unsupervised}   & 23 & 53.6K  & 18 \\
    UAVDark135~\cite{li2022all}                         & 19 & 41.7K  & 15 \\
    \midrule
    \cellcolor{gray!11}NAT2024-1      & \cellcolor{gray!11}\textbf{40} & \cellcolor{gray!11}\textbf{70.0K}  & \cellcolor{gray!11}\textbf{35} \\
    \bottomrule
    \end{tabular}
    }
  \label{tab:dataset}
\end{table}

\section{Experiments}
%%%%%%%%%%%%%%%%%%%%%%%%%%%%%
In this section, the detailed implementation is provided. Long-term tracking performance evaluation and illumination-oriented analysis demonstrate that TDA-Track surpasses other lightweight trackers in long-term nighttime UAV tracking.
Moreover, an ablation study is introduced to testify the effectiveness of the proposed temporal domain adaptation framework. Finally, real-world tests validate the remarkable performance of TDA-Track in UAV applications.

\begin{figure*}[t]	
    \centering
    \vspace{10pt}
    \colorbox{figure_c}{
        \includegraphics[width=0.31\linewidth]{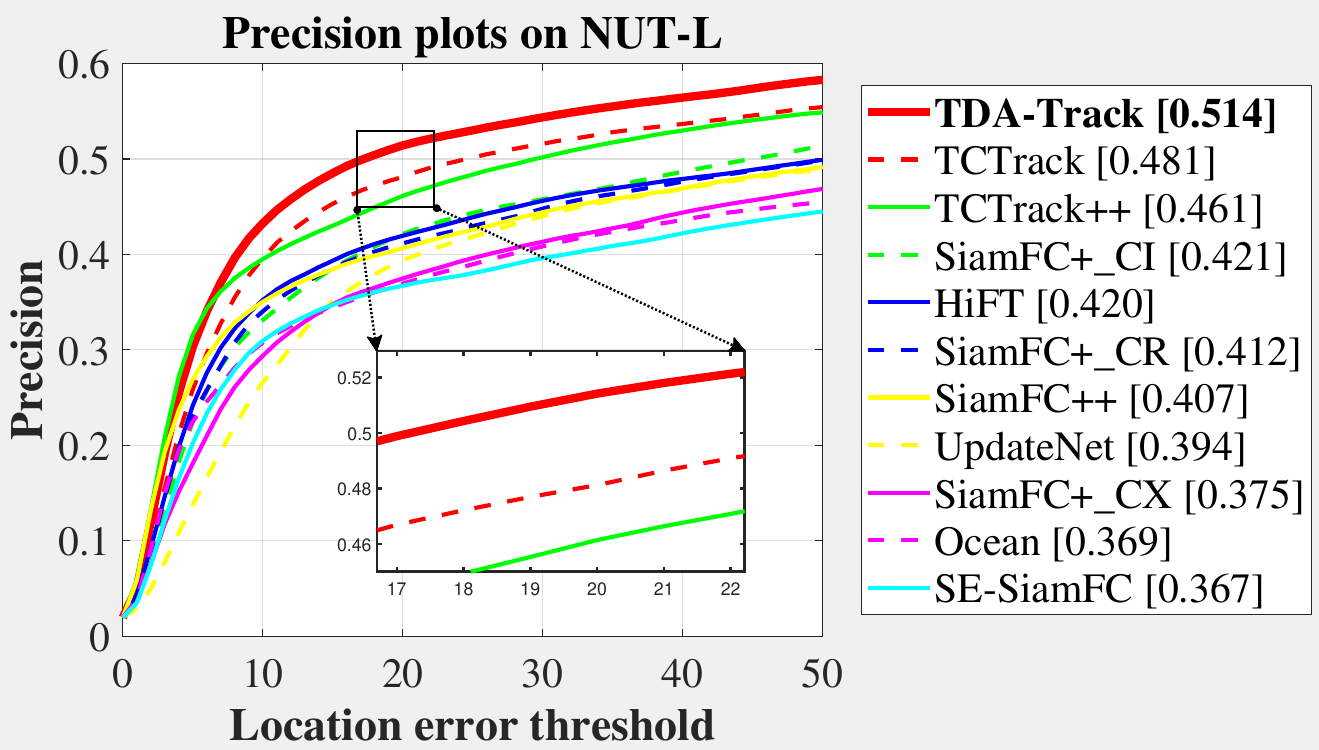}
        \includegraphics[width=0.31\linewidth]{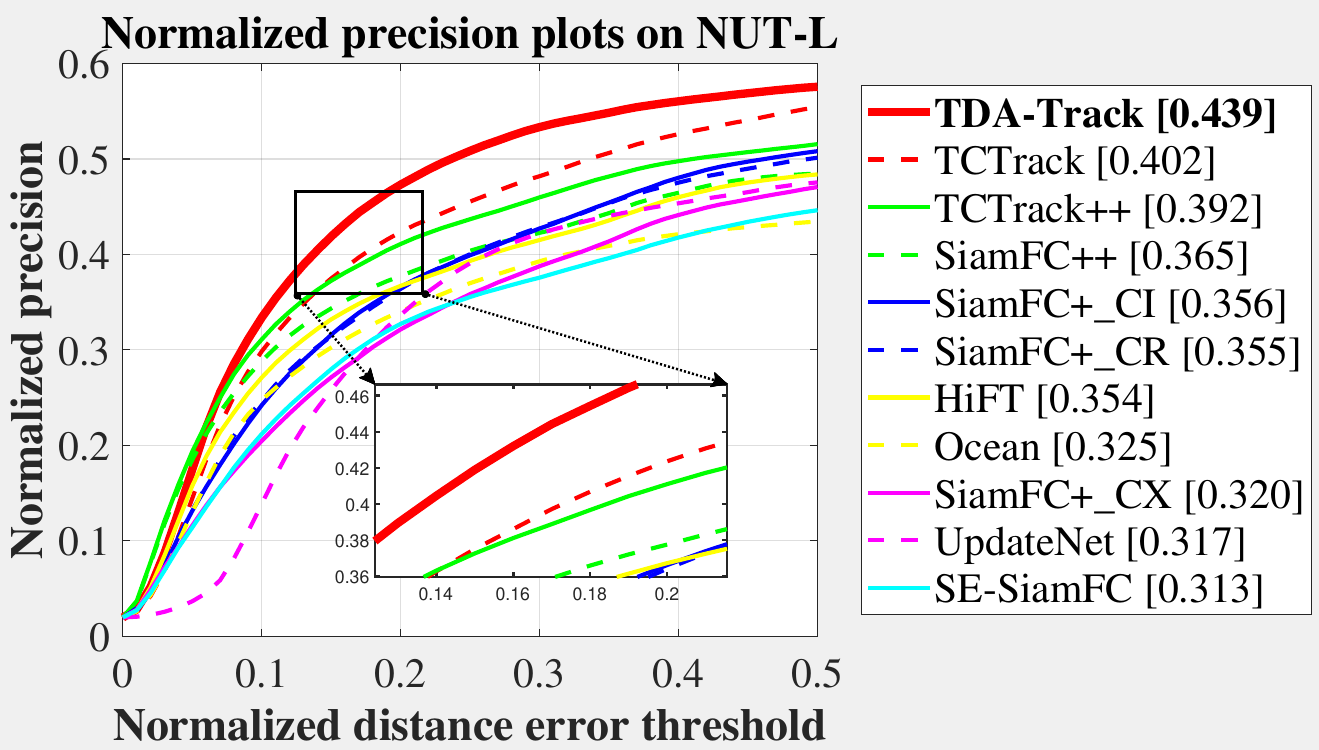}
        \includegraphics[width=0.31\linewidth]{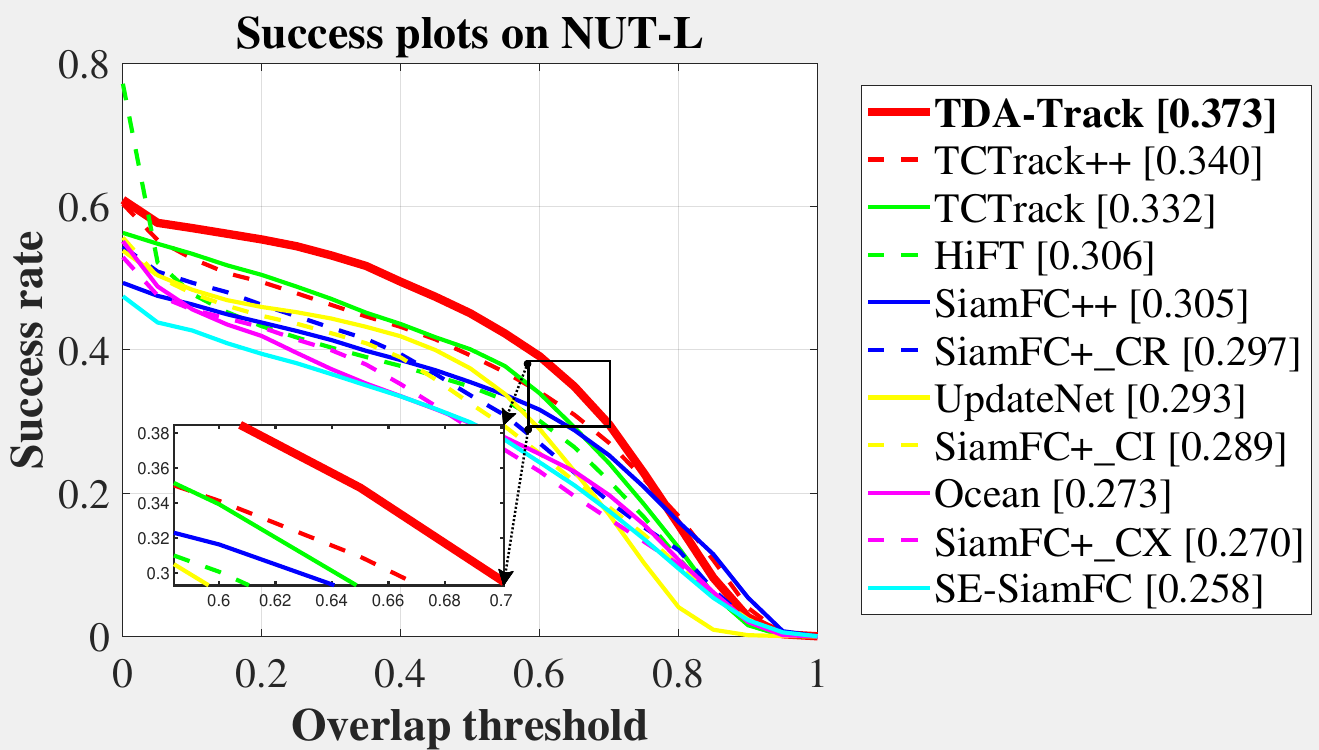}
        \setlength{\abovecaptionskip}{4pt}
        }
    \colorbox{figure_c}{
        \includegraphics[width=0.31\linewidth]{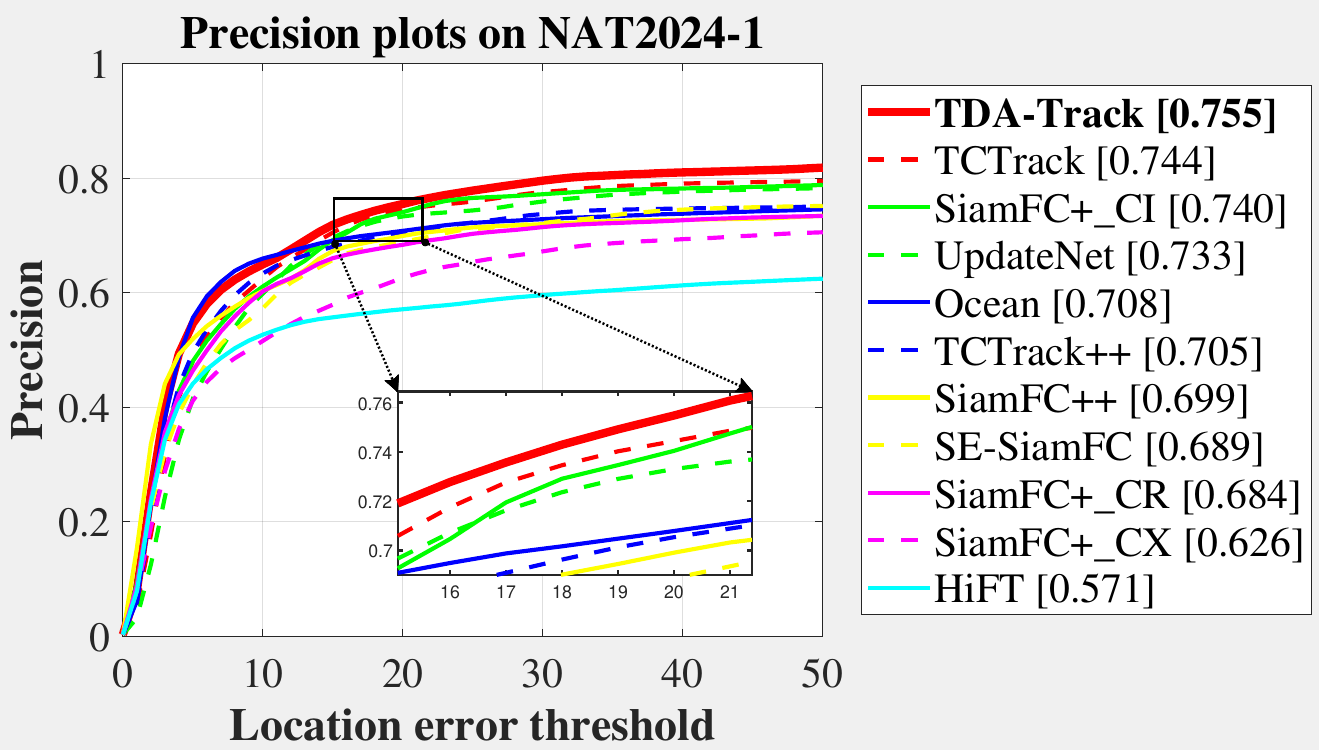}
        \includegraphics[width=0.31\linewidth]{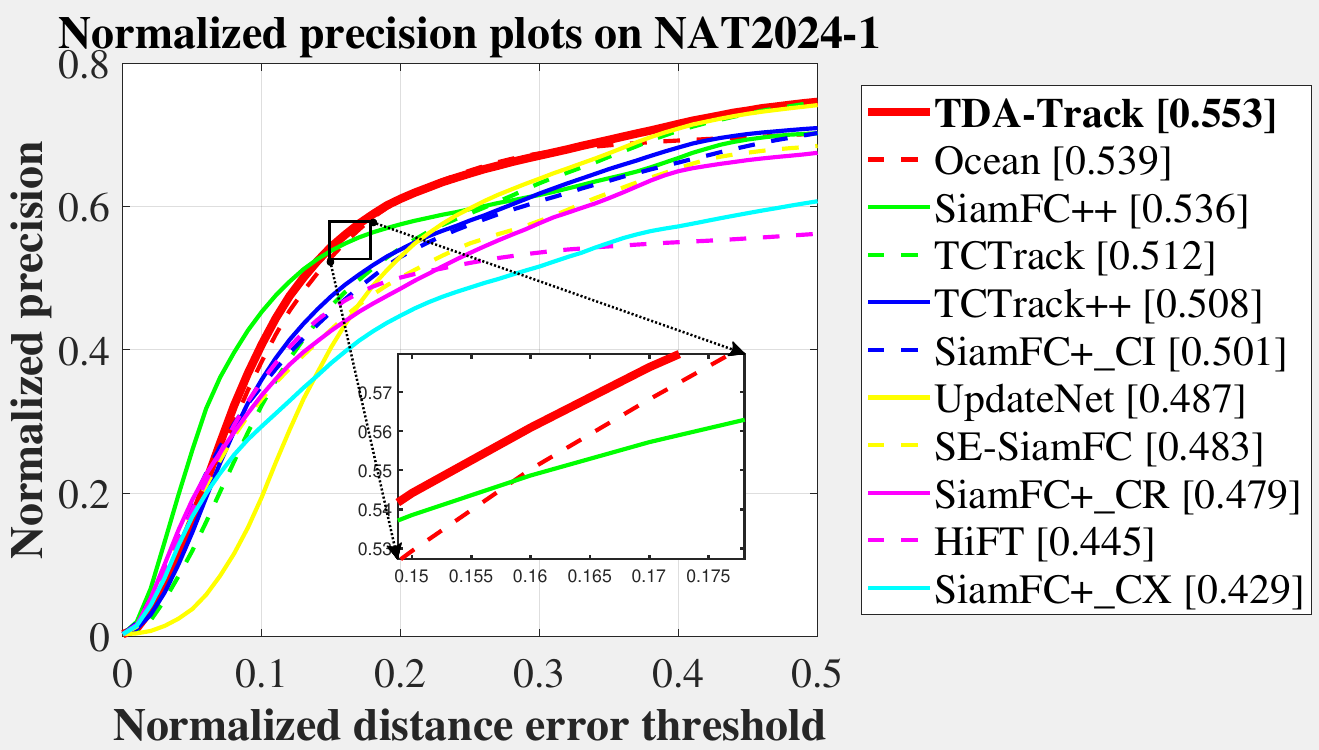}
        \includegraphics[width=0.31\linewidth]{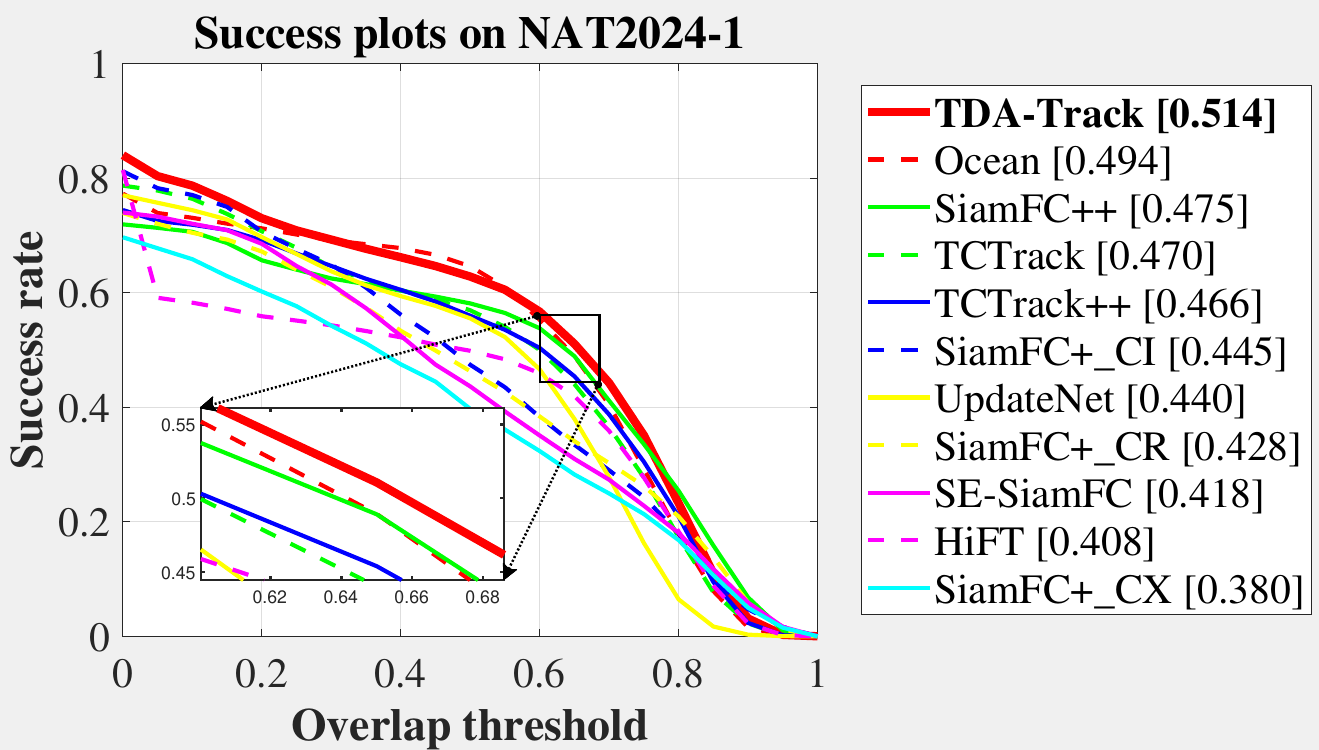}
        \setlength{\abovecaptionskip}{4pt}
        }
        %colorbox{figure_c}{}

    \caption{Long-term nighttime UAV tracking performance of TDA-Track and lightweight trackers on NUT-L and NAT2024-1. TDA-Track ranks first in all three metrics with remarkable improvement.}
    \setlength{\abovecaptionskip}{4pt}
    \vspace{-10pt}
    \label{fig:fig5 offline performance}
    % \vspace{-15pt}
\end{figure*}	

%%%%%%%%%%%%%%%%%%%%%%%%%%%%%
\subsection{Implementation Details}
The proposed TDA training framework is implemented using PyTorch and is trained for 25 epochs on an A100 GPU. In terms of data preparation, the daytime training samples are extracted from ImageNet VID~\cite{russakovsky2015imagenet} and GOT-10K~\cite{huang2019got}, while the nighttime ones are cropped from NAT2021-\textit{train}~\cite{ye2022unsupervised}. The multi-modal tracking model~\cite{nettrack} is used to implement the prompt-driven object mining approach. For faster training, TDA-Track is initialized with pre-trained parameters of its Baseline tracker~\cite{cao2022tctrack}. 
%The learning rate warms up from \(3 \times 10^{-4}\) to \(1.5 \times 10^{-3}\), and subsequently decays to \(1.5 \times 10^{-5}\) in log space. 
Note that the backbone parameters in TDA-Track are frozen for the first 10 epochs. Moreover, the temporal-consistent Transformer discriminator is optimized by Adam Optimizer~\cite{kingma2014adam} with a base learning rate of 0.005 and poly learning rate policy.
\subsection{Evaluation Metrics}
The one-pass evaluation (OPE)~\cite{mueller2016benchmark} is adopted to rank tracking performances with precision, normalized precision, and success rate. The precision is calculated by the Euclidean distance between the ground truth centers and predicted bounding box centers, \textit{i.e.}, the center location error (CLE). The precision plot demonstrates the percentage of frames where CLE is less than the threshold. The normalized precision plot normalizes precision with ground truth sizes to alleviate the influence of object sizes. The success rate plot computes the percentage of frames where the intersection over union (IoU) of the ground truth and predicted bounding box is greater than a given threshold.
%%%%%%%%%%%%%%
\subsection{Overall Performance}
TDA-Track has a hybrid architecture with a lightweight CNN backbone and an efficient Transformer neck. Hence it's compared with lightweight CNN trackers~\cite{xu2020siamfc++, zhang2019learning, zhang2020ocean, sosnovik2021scale, Zhang2019SiamDW} and efficient CNN-Transformer hybrid trackers~\cite{cao2021hift, cao2022tctrack,cao2023towards}. 
The evaluation is conducted on long-term tracking benchmarks to rate the power of temporal contexts for nighttime UAV tracking. The results are shown in Fig.~\ref{fig:fig5 offline performance}.

\begin{figure}[!t]
    \centering
    \vspace{-5pt}
    \colorbox{figure_c}{
        \includegraphics[width=0.93\linewidth]{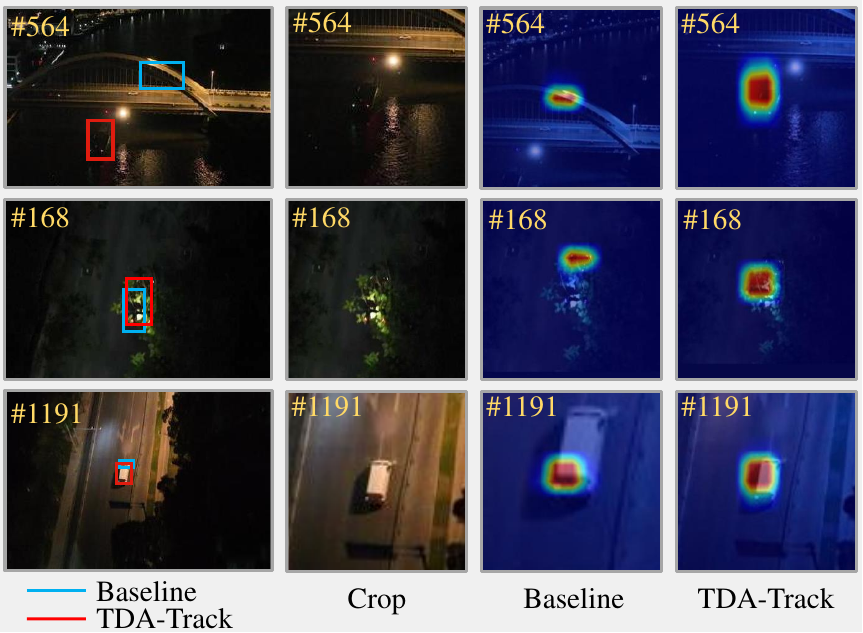}
        }
    \caption{Visualization of the confidence map and tracking performance on several sequences. The first row showcases that TDA-Track (\textcolor{red}{red}) tracks the ship robustly despite the severe low ambient intensity (LAI). More cases prove that TDA-Track achieves better performance than Baseline tracker~\cite{cao2022tctrack} (\textcolor{blue}{blue}) at nighttime.}
    \label{fig:heatmap}
    \vspace{-15pt}
\end{figure}

\indent \textbf{NUT-L}: 
NUT-L~\cite{fu2024SAMDA} is a long-term nighttime UAV tracking benchmark collecting the long sequences of NAT2021-\textit{L-test}~\cite{ye2022unsupervised} and UAVDark135~\cite{li2022all}. TDA-Track surpasses the Baseline tracker~\cite{cao2022tctrack} by a large margin and achieves first rank in precision (\textbf{0.514}), normalized precision(\textbf{0.439}), and success rate (\textbf{0.373}), which makes \textbf{6.9\%}, \textbf{9.2\%}, and \textbf{12.3\%} improvement over Baseline tracker~\cite{cao2022tctrack} on the three metrics respectively. \\
\indent \textbf{NAT2024-1}: TDA-Track wins first price in precision (\textbf{0.755}), normalized precision(\textbf{0.553}), and success rate (\textbf{0.514}), which achieves \textbf{0.011}, \textbf{0.041}, \textbf{0.044} gains in three metrics compared with Baseline tracker~\cite{cao2022tctrack}.\\
\indent As shown in Fig.~\ref{fig:heatmap}, several confidence maps of TDA-Track and Baseline tracker~\cite{cao2022tctrack} are visualized using Grad-Cam~\cite{selvaraju2017grad}. While the Baseline tracker has difficulty concentrating on objects at nighttime, TDA-Track has stronger nighttime perception ability and boosts nighttime tracking performance.\\
\Remark The proposed framework brings about favorable performance improvement on the benchmarks. The results prove the generalization capability and application potential of TDA-Track in various nighttime conditions.
%%%%%%%%%%%%%%%%%%%%%%%%%%%%%
\subsection{Illumination-Oriented Evaluation}
\begin{table}[!b]
  \centering
  \vspace{-5pt}
  \caption{Illumination-oriented evaluation of lightweight trackers. The top 2 performances are highlighted in \textcolor{red}{red}, and \textcolor{blue}{blue}.}
  \vspace{-5pt}
  \colorbox{table_c}{
  \resizebox{0.95\linewidth}{!}{
  %\setlength{\heavyrulewidth}{1.2pt}
    %\begin{tabular}{c|cc|cc|}
    \begin{tabular}{>{\centering\arraybackslash}p{2.4cm}|p{1cm}p{1cm}|p{1cm}p{1cm}}
        \toprule
        \multirow{2}{*}{\diagbox[height=3em, width=2.7cm]{Trackers}{Challenges}}&\multicolumn{2}{c|}{LAI} &\multicolumn{2}{c}{IV}\\
        \cmidrule(lr){2-3} \cmidrule(lr){4-5}
        %&
        %\multicolumn{2}{c|}{low ambient intensity}&\multicolumn{2}{c|}{low ambient intensity}\\
        %\cmidrule(lr){2-3} \cmidrule(lr){4-5} 
        & 
        \multicolumn{1}{>{\centering\arraybackslash}m{1.cm}}{Prec.} & \multicolumn{1}{>{\centering\arraybackslash}m{1.cm}|}{Succ.} & \multicolumn{1}{>{\centering\arraybackslash}m{1.cm}}{Prec.} & \multicolumn{1}{>{\centering\arraybackslash}m{1.cm}}{Succ.}\\
        \midrule
        SiamFC+\_CX \cite{Zhang2019SiamDW}  & \centering0.347 & \centering0.254 & \centering0.351 & \centering\arraybackslash0.239 \\
        SiamFC+\_CI \cite{Zhang2019SiamDW}  & \centering0.385 & \centering0.269 & \centering0.379 & \centering\arraybackslash0.248\\
        SiamFC+\_CR \cite{Zhang2019SiamDW}  & \centering0.376 & \centering0.287 & \centering0.362 & \centering\arraybackslash0.246\\
        UpdateNet \cite{zhang2019learning}  & \centering0.396 & \centering0.309 & \centering0.351 & \centering\arraybackslash0.262\\
        Ocean~\cite{zhang2020ocean}         & \centering0.313 & \centering0.238 & \centering0.318 & \centering\arraybackslash0.237\\
        SiamFC++~\cite{xu2020siamfc++}      & \centering0.365 & \centering0.275 & \centering0.410 & \centering\arraybackslash\textcolor{blue}{0.307}\\
        SE-SiamFC~\cite{sosnovik2021scale}  & \centering0.328 & \centering0.232 & \centering0.325 & \centering\arraybackslash0.217\\
        HiFT \cite{cao2021hift}             & \centering0.391 & \centering0.290 & \centering0.389 & \centering\arraybackslash0.276\\
        TCTrack \cite{cao2022tctrack}       & \centering\textcolor{blue}{0.450} & \centering\textcolor{blue}{0.319} & \centering\textcolor{blue}{0.446} & \centering\arraybackslash0.304\\
        TCTrack++ \cite{cao2023towards}     & \centering0.431 & \centering0.313& \centering0.411 & \centering\arraybackslash0.295\\
        \midrule
        \textbf{TDA-Track}                  & \centering\textbf{\textcolor{red}{0.512}}& \centering\textbf{\textcolor{red}{0.375}}& \centering\textbf{\textcolor{red}{0.467}} & \centering\arraybackslash\textbf{\textcolor{red}{0.334}}\\
        \bottomrule
    \end{tabular}
    \label{tab:attribute}
    }
    }
\end{table}
\begin{table}[!b]
  \centering
  \vspace{-10pt} % Reduce the space above the table
  \caption{Ablation study of the proposed methodology on NUT-L.}
  \vspace{-7pt} % Reduce the space between caption and table
  \colorbox{table_c}{
  \resizebox{0.95\linewidth}{!}{
    \begin{tabular}{ccccccc}
    \toprule
    \multirow{2}{*}{Trackers} & \multicolumn{2}{c}{Align} & \multicolumn{2}{c}{Disc.} &\multirow{2}{*}{Norm.Prec.} & \multirow{2}{*}{Succ.}\\ 
    \cmidrule(r){2-3} \cmidrule(r){4-5}
             & IF & TC   & PD   & TCD    &    &      \\ 
    \midrule
    Baseline tracker         &    &       &     &       & 0.402 & 0.332 \\
    Baseline tracker+OM      &\cmark    &       &\cmark     &       & 0.421 & 0.357 \\
    Baseline tracker+OM+TC   &\cmark    &\cmark &\cmark     &       & 0.419 & 0.357 \\
    \midrule
    \cellcolor{gray!10}\textbf{TDA-Track}  &\cmark &\cmark & & \cmark & \cellcolor{gray!10}\textbf{0.439} &\cellcolor{gray!10}\textbf{0.373}\\
    \bottomrule
    \end{tabular}
    }
    }
  \label{tab: ablation}
\end{table}
The discrepancy between daytime and nighttime domains is mainly caused by different illumination conditions. To investigate the performance of TDA-Track at nighttime, we perform an analysis oriented for the low ambient illumination (LAI) and the illumination variation (IV) challenge on NUT-L. As shown in TABLE~\ref{tab:attribute}, concerning the LAI attribute, TDA-Track achieves remarkable \textbf{13.8\%} and \textbf{17.6\%} improvement in precision and success rate compared to Baseline tracker~\cite{cao2022tctrack}. Moreover, TDA-Track is far ahead of other light-weight trackers on IV challenge in precision(\textbf{0.467}) and success rate (\textbf{0.334}).
%%%%%%%%%%%%%%%%%%%%%%%%%%%%%
\subsection{Ablation Study}
In this section, contributions of the proposed methodology are verified on NUT-L, as demonstrated in TABLE~\ref{tab: ablation}. To prove the superiority of prompt-driven object mining, domain adaptation training on the uncovered samples (Baseline tracker+OM) aligns image features (IF) by learning against the plain discriminator (PD)~\cite{ye2022unsupervised}, which achieves \textbf{4.7\%} and \textbf{7.5\%} improvement in normalized precision and success rate compared to Baseline tracker~\cite{cao2022tctrack}. Then to introduce temporal context (TC) into framework, the UAV tracker is trained to learn domain-invariant temporal contexts (Baseline tracker+OM+TC) against PD, but fails to achieve substantial further performance improvement due to incoherent discrimination results of PD. Finally, the proposed TDA framework replaces PD with the temporal-consistent discriminator (TCD) and exploits full potential of temporal contexts. TDA-Track achieves favorable nighttime tracking performance with normalized precision (\textbf{0.439}) and success rate (\textbf{0.373}).
\vspace{-5pt}
%%%%%%%%%%%%%%%%%%%%%%%%%%%%%%%%%%%%%%%%%%%%%%%%%%%%%%%%%%%
%%%%%%%%%%%%%%%%%%%%%%%%%%%%%%%%%%%%%%%%%%%%%%%%%%%%%%%%%%%
\section {Real-World Tests}
\begin{figure}[!t]
    \centering
    \colorbox{figure_c}{
        \includegraphics[width=0.93\linewidth]{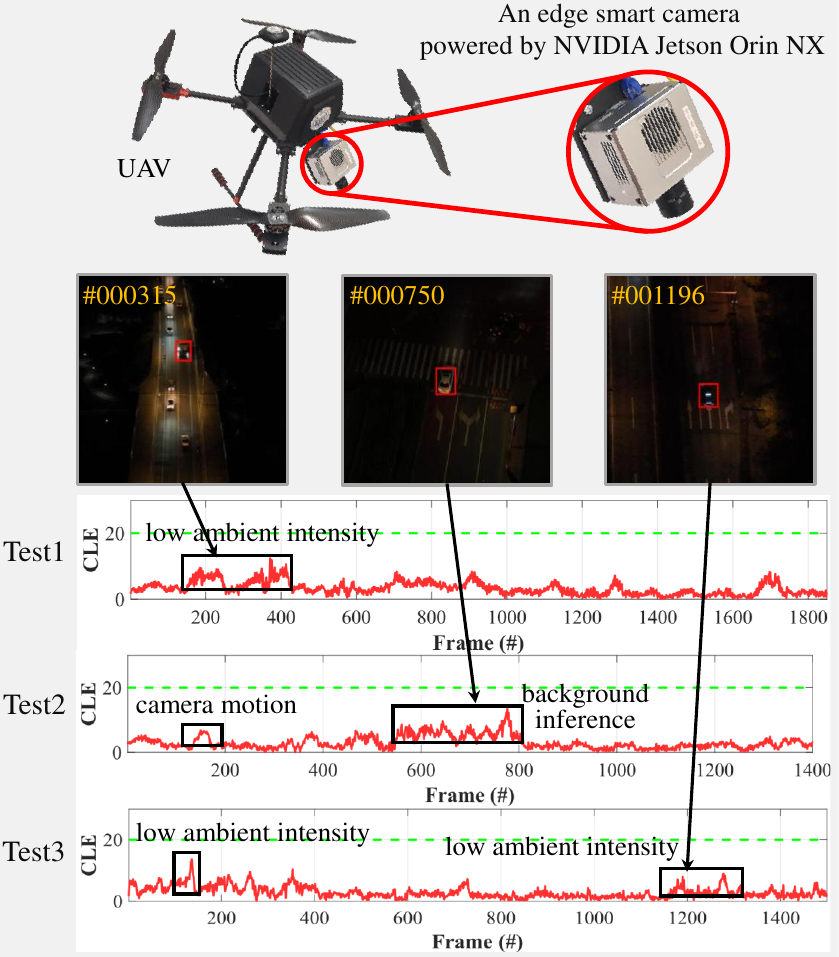}
    }
    \caption{Real-world tests on UAV platform equipped with NVIDIA Jetson Orin NX prove the robustness and precision of TDA-Track. The \textcolor{red}{red} bounding boxes represent the tracking results.}
    \label{fig:realworldtest}
    \vspace{-15pt}
\end{figure}

Extensive real-world experiments are conducted on a UAV platform equipped with an edge smart camera powered by NVIDIA Jetson Orin NX, as shown in Fig.~\ref{fig:realworldtest}, to evaluate real-world performance. Several challenging long sequences in unfavorably illuminated scenarios are presented. In Test 1, the drone captures the scene from a high altitude, where the truck intermittently drives into the darkness. In Test 2, the drone follows the car in a horizontal flight for a considerable distance. The camera motion and background inference pose great challenges to tracking. In Test 3, the low ambient intensity exacerbates the tracking difficulty. To assess tracking performance, the center location error (CLE) is adopted as the criteria. Despite the challenges of the three sequences, TDA-Track tracks the objects robustly. In addition, TDA-Track is real-time running with 32 frames per second. We are confirmed that TDA-Track is remarkable in both low latency and strong long-term nighttime tracking performance, which is suitable for edge deployment on UAV platforms.
%%%%%%%%%%%%%%%%%%%%%%%%%%%%%%%%%%%%%%%%%%%%%%%%%%%%%%%%%%%
%%%%%%%%%%%%%%%%%%%%%%%%%%%%%%%%%%%%%%%%%%%%%%%%%%%%%%%%%%%
\section{Conclusions}
In this work, a temporal DA training framework for nighttime UAV tracking, namely TDA, is developed. The temporal generator learns to coherently map consecutive frames into domain-invariant temporal contexts and image features by adversarial training. The domain discriminator is designed to extract common representations from encoded features, which enables more accurate domain classification and better domain adaptability. Furthermore, a prompt-driven object mining method obtains high-quality training samples of valuable objects in smooth trajectories. Moreover, a long-term nighttime UAV tracking benchmark, namely, NAT2024-1 is constructed. Evaluation of TDA-Track, the corresponding tracker trained under TDA framework, proves the effectiveness of our framework in boosting nighttime tracking performance. We're convinced that the temporal domain adaptation framework can boost nighttime UAV tracking with temporal context alignment.
%%%%%%%%%%%%%%%%%%%%%%%%%%%%%%%%%%%%%%%%%%%%%%%%%%%%%%%%%%%
%%%%%%%%%%%%%%%%%%%%%%%%%%%%%%%%%%%%%%%%%%%%%%%%%%%%%%%%%%%
\section*{Acknowledgement}
This work is supported by the National Natural Science Foundation of China (No. 62173249) and the Natural Science Foundation of Shanghai (No. 20ZR1460100).

\bibliographystyle{IEEEtran}
\bibliography{IEEEabrv,ref} %the second parameter is name of reference bib file
\end{document}